\documentclass[letterpaper, 10 pt, conference]{ieeeconf}
\IEEEoverridecommandlockouts   
\usepackage{times}
\usepackage{url}
\usepackage{xcolor}
\usepackage{graphicx}
\usepackage{glossaries} 
\usepackage [utf8]{inputenc}
\usepackage[sorting=none]{biblatex}

\def\BibTeX{{\rm B\kern-.05em{\sc i\kern-.025em b}\kern-.08em
    T\kern-.1667em\lower.7ex\hbox{E}\kern-.125emX}}
\begin{document}

\newacronym{batman}{B.A.T.M.A.N}{Better Approach To Mobile Ad-hoc Networking }
\newacronym{dds}{DDS}{Data Distribution Service}
\newacronym{dymu}{DyMu}{Dynamic-Multi-Layered Path Planning}
\newacronym{esa}{ESA}{European Space Agency}
\newacronym{esric}{ESRIC}{European Space Resources Innovation Centre}
\newacronym{fps}{fps}{frames per second}
\newacronym{gui}{GUI}{graphical user interface}
\newacronym{imu}{IMU}{Inertial Measurement Unit}
\newacronym{isru}{ISRU}{In-Situ Resources Utilisation}
\newacronym{lidar}{LiDAR}{Light Detecting And Ranging}
\newacronym{lro}{LRO}{Lunar Reconnaissance Orbiter}
\newacronym{mrs}{MRS}{Multi-Robot Systems}
\newacronym{realms}{REALMS}{Resilient Exploration And Lunar Mapping System}
\newacronym{realms2}{REALMS2}{Resilient Exploration And Lunar Mapping System 2}
\newacronym{rgbd}{RGB-D}{RGB-Depth}
\newacronym{ros}{ROS}{Robot Operating System}
\newacronym{ros2}{ROS 2}{Robot Operating System version 2}
\newacronym{slam}{SLAM}{Simultaneous Localisation And Mapping}
\newacronym{snt}{SnT}{Centre for Security, Reliability and Trust}
\newacronym{roi}{ROI}{region of interest}
\newacronym{rtabmap}{RTAB-Map}{Real-Time Appearance Based Mapping}
\newacronym{viper}{VIPER}{Volatiles Investigating Polar Exploration Rover}
\newacronym{vslam}{vSLAM}{Visual Simultaneous Localisation And Mapping}
\newacronym{v&v}{V\&V}{Verification and Validation}
\newacronym{hwmp}{HWMP}{Hybrid Wireless Mesh Protocol}
\newacronym{sas}{SAS}{Space Applications Services}
\newacronym{manet}{MANET}{Mobile Ad-Hoc Networks}
\newacronym{per}{PER}{Packet Error Rate}
\newacronym{cots}{COTS}{Commercial-of-the-Shelf}
\newacronym{los}{LOS}{line-of-sight}


\title{\LARGE \bf REALMS2 - Resilient Exploration And Lunar Mapping System 2 – A Comprehensive Approach
}

\author{Dave van der Meer$^{1}$$^{*}$, Loïck P. Chovet$^{1}$$^{*}$, Gabriel M. Garcia$^{1}$$^{*}$,\\ Abhishek Bera$^{1}$ and Miguel Angel Olivares-Mendez$^{1}$
\thanks{This research was funded in whole, or in part, by the Luxembourg National Research Fund (FNR), grant references 14783405, 17025341 and 17679211. For the purpose of open access, and in fulfilment of the obligations arising from the grant agreement, the author has applied a Creative Commons Attribution 4.0 International (CC BY 4.0) license to any Author Accepted Manuscript version arising from this submission. This project was partly funded by the ESA-ESRIC Space Resources Challenge, Contract No. 4000137334/22/NL/AT.}%
\thanks{$^{1}$Affiliated at  Space Robotics (SpaceR) Research Group, Interdisciplinary Centre for Security, Reliability and Trust (SnT), University of Luxembourg, Luxembourg, Luxembourg {\tt \{dave.vandermeer}, {\tt loick.chovet}, {\tt gabriel.garcia}, {\tt abhishek.bera}, {\tt miguel.olivaresmendez\}@uni.lu}}
\thanks{* Authors share equal contribution.}
}

\maketitle
\thispagestyle{empty}
\pagestyle{empty}

\begin{abstract}
The \gls{esa} and the \gls{esric} created the Space Resources Challenge to invite researchers and companies to propose innovative solutions for \gls{mrs} space prospection.
This paper proposes the \gls{realms2}, a \gls{mrs} framework for planetary prospection and mapping.
Based on \gls{ros2} and enhanced with \gls{vslam} for map generation, \gls{realms2} uses a mesh network for a robust ad hoc network.
A single \gls{gui} controls all the rovers, providing a simple overview of the robotic mission.
This system is designed for heterogeneous multi-robot exploratory missions, tackling the challenges presented by extraterrestrial environments.
\gls{realms2} was used during the second field test of the \gls{esa}-\gls{esric} Challenge and allowed to map around 60\% of the area, using three homogeneous rovers while handling communication delays and blackouts.
\end{abstract}

\glsresetall


\section{Introduction}
\label{introduction}

\begin{figure} [htp]
\centering
\includegraphics[width=0.8\columnwidth]{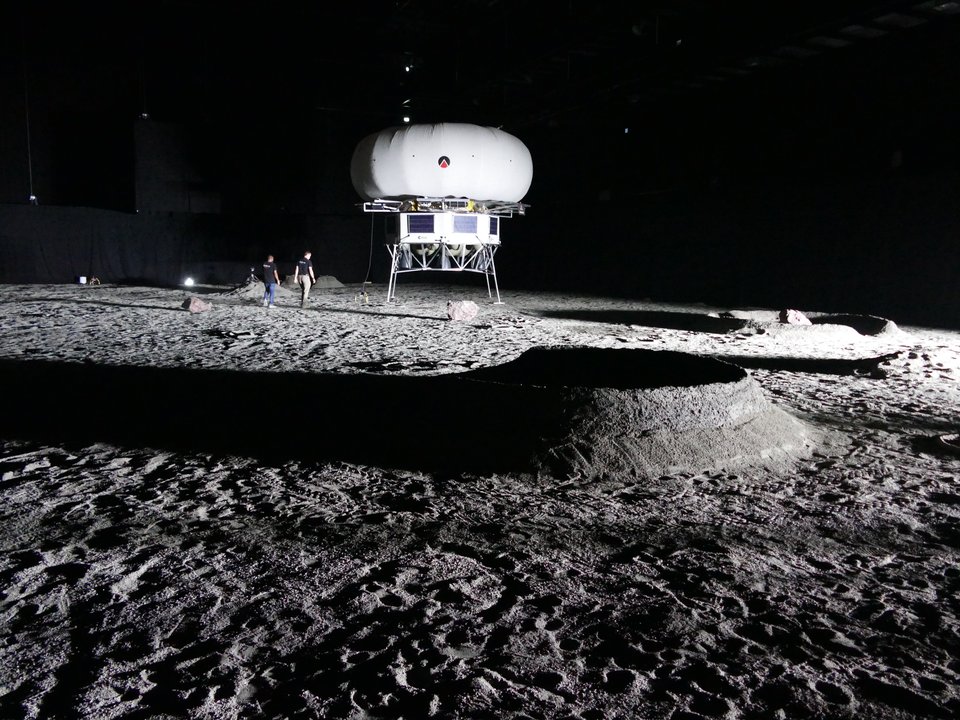}
\caption{Second field test arena for the ESA-ESRIC Space Resources Challenge}
\label{fig:ESA_ESRIC_SRC_second_field_test_arena}
\end{figure}

Recently, the Moon has regained the focus of space agencies and private companies for potential \gls{isru}.
Therefore, the \gls{esa} and the \gls{esric} seek to increase the level of autonomy of robotic systems used for the exploration of space resources.
\gls{esa} and \gls{esric} organised the Space Resources Challenge~\cite{esa_esric_src_2023a}, where 13 research teams competed in a first field test to demonstrate their concepts of autonomous systems, leveraging the advantages of \gls{mrs}.
The five best teams continued to a second field test~\cite{esa_esric_src_2023b} with the task of finding different resources within a large lunar analogue environment, shown in Fig~\ref{fig:ESA_ESRIC_SRC_second_field_test_arena}.

During the first field test of the Challenge~\cite{esa_esric_src_2023b}, the authors present the \gls{realms}~\cite{REALMS1_2023}, a \gls{mrs} using two rovers mapping the environment with \gls{vslam}.
\gls{realms} is based on the \gls{ros} to communicate with other nodes on the robots.
The first part of the challenge revealed multiple limitations of \gls{realms}~\cite{REALMS1_2023}.
First, the \gls{ros} framework is designed for centralised systems and provides only limited support for independent \gls{mrs}. Second, the robustness of the communication system needs to be improved. Third, \gls{realms} offers limited scalability due to the design of the user interface and due to the high network load that severely limits the downlink budgets.

This paper proposes \gls{realms2}, an improved version of \gls{realms} that addresses these limitations.
The revised architecture has increased coverage and resilience based on the \gls{ros2} framework that provides improved \gls{mrs} support, making each robot a fully standalone system.

The contributions of this work are:
\begin{itemize}
    \item A fully decentralized and scalable \gls{mrs} architecture built on \gls{ros2}, enabling modular deployment, fault tolerance, and seamless multi-robot coordination in communication-constrained lunar-like environments.
    
    \item A mesh-based communication layer using HWMP+, validated in integration with \gls{ros2}, allowing dynamic relay formation and resilience to network degradation or loss.
    
    \item A complete field-validated system (\gls{realms2}) tested during the \gls{esa}-\gls{esric} Challenge, including live map merging capabilities, bandwidth-aware robot control, and single-operator mission management under simulated lunar constraints.
\end{itemize}

The paper is structured as follows:
Section \ref{sec:related_work} highlights the state-of-the-art in planetary robotics.
Section \ref{sec:methods} briefly shows the system setup.
Section \ref{sec:experiments} describes the system validation through a series of experiments.
Section \ref{sec:challenge} shows the use case during the \gls{esa}-\gls{esric} Space Resources Challenge.
Section \ref{sec:results} presents the results of the experiments and our participation in the challenge.
Sections \ref{sec:discussion} and \ref{sec:conclusion} highlight the lessons learned from \gls{realms2} and its application in the challenge.


\color{black}
\section{Related Work}
\label{sec:related_work}

Most planetary robotics missions consisted of a lander and a single robot. However, new approaches highlight \gls*{mrs} as the correct approach to ensure resilience, range, and cost efficiency as shown by recent projects such as NASA's CADRE \footnote{https://www.jpl.nasa.gov/missions/cadre} or Ingenuity \footnote{The ingenuity helicopter on the perseverance rover} with Perseverance mission.

This section highlights the existing work in the field of \gls{mrs} along with space-related applications. It also details the Mesh Network technology and \gls{ros2} used for \gls*{mrs} communication

\subsection{Multi-robot systems}

The authors of \cite{Parker_2008} distinguished four \gls{mrs} architectures:

\begin{itemize}
    \item Centralised: A single central node is in charge of operating all the robots. It offers a more optimised approach since all the knowledge is centralised, but it can lead to network bottlenecks, and any malfunction on the central node can lead to an overall failure.
    \item Hierarchical: This approach features smaller centralised systems coordinated by bigger central nodes in a hierarchical way. This allows for more stability toward failure since each central node maintains the capability to operate its fleet.
    \item Decentralised: Each robot is in charge of making its own decisions, depending on the data transmitted by the other robots. It is the most robust approach; however, it leads to suboptimal results and might lead to more network load because all robots need to gather data from the rest of the system.
    \item Hybrid: This approach combines the previous methods to achieve robustness with efficiency.
\end{itemize}

Space missions have unique characteristics that should impact the choice of architecture.
In the case of a centralised architecture, a master node is required to handle all robot tasks and to respond quickly to emergencies.
However, there is a delay in communication between the Earth and any astronomical object.
This delay would force the master node to be hosted directly on the same celestial body.
The key to success in space missions is redundancy, and having one master node would lead to a single point of failure, potentially affecting the whole system.
According to these rules, the architecture targeted for space \gls{mrs} should be at least partially a decentralised architecture, allowing enough redundancy and resilience. 
With this logic in mind, the LUNARES \cite{cordes_lunares_2011} project was presented.
The researchers implemented a team of heterogeneous robots in charge of autonomous data sampling.
One of the key elements of the system is the Ground Control Station, in charge of supervising critical operations.
This study mainly demonstrates the capabilities of each robot, along with their ability to collaborate. ARCHES \cite{schuster_arches_2020}, which focused on the exploration of lava tubes, presented a more recent approach.
ARCHES featured a drone for the first time.
It demonstrated the growing interest in decentralised heterogeneous \gls{mrs} for space applications. 

\subsection{Mesh Networks}

Mesh networks represent a network topology in which every node can act as a relay, further propagating the network.
It was introduced in the IEEE 802.11s norm \cite{hiertz_ieee_2010} as a new topological approach to wireless networks.
This approach to networking has often been shown to be the best solution for \gls{manet} \cite{akyildiz_wireless_2005}.
It offers the following advantages :

\begin{itemize}
    \item \textbf{Robustness}: Since every robot is creating and propagating the network, if any robot fails, rerouting is dynamically triggered.
    \item \textbf{Scaling}: Any additional node can be dynamically added to the system. This can lead to a better \textbf{coverage} with the addition of relay nodes
    \item \textbf{Redundancy}: Multiple robots can ensure a link. It ensures fail-safe behaviour, which is needed for any space mission.
\end{itemize}

The IEEE 802.11s norm \cite{802_11s_2011} also introduces a mesh networking protocol called \gls{hwmp}.
It introduces a metric called airtime, which is used to define the best way to communicate messages.
The authors of \cite{yang_hwmp_2012} studied issues related to \gls{hwmp}.
They highlighted that the airtime metric did not take into account traffic flow, leading to suboptimal routing.
They also show that \gls{hwmp} leads to more data overhead.
To solve these issues, they introduce \gls{hwmp}+, implementing two main improvements.
They change the way the \gls{per} is computed to get a better estimation of the link quality.
The second change is an observation of the traffic load on the link, impacting the computation of the metric.

Other approaches to mesh networking exist, such as AODV \cite{perkins_ad_2003} and BATMAN \cite{jafri_split_2023}.
These approaches have some advantages; For instance, they are compatible with any wireless device, in contrast to \gls*{hwmp}, which must adhere to IEEE 802.11s standards. 
However, these approaches also feature issues, such as the routing protocol not estimating the link quality.


\section{Methods and Materials}
\label{sec:methods}

We present \gls{realms2} as a scalable \gls{mrs} for space exploration.
It features three main subsystems: the rovers, the lander, and the ground station for the operators.
Each operator can access the interface to send goals to each of the robots to explore the surroundings.
The system is designed to address the challenges of a lunar mission, such as communication delays and the environment.

Compared to REALMS, the rovers have a more robust mapping system, more robust headlight controls, more processing power, and extended battery life.
The user interface allows monitoring and controlling each robot through a single interface, making the control system more scalable and easier for a single operator to use.
The upgrade to \gls{ros2} allows easier integration of additional robots and simplifies the communication protocol for \gls{mrs}.
The introduction of a lunar lander adds a central interface between the rovers and the ground station, as would be the case for a real lunar mission.
The lander also acts as an additional resource for offloading high-computational processing and serves as a host for the sensors that overview the close environment and the rovers' departure.

\subsection{Rover architecture}

The \gls{realms2} rovers consist of a variety of hardware and software components chosen for this specific mission of lunar exploration.

\subsubsection{Hardware}

\begin{figure} [htp]
\centering
\includegraphics[width=0.9\columnwidth]{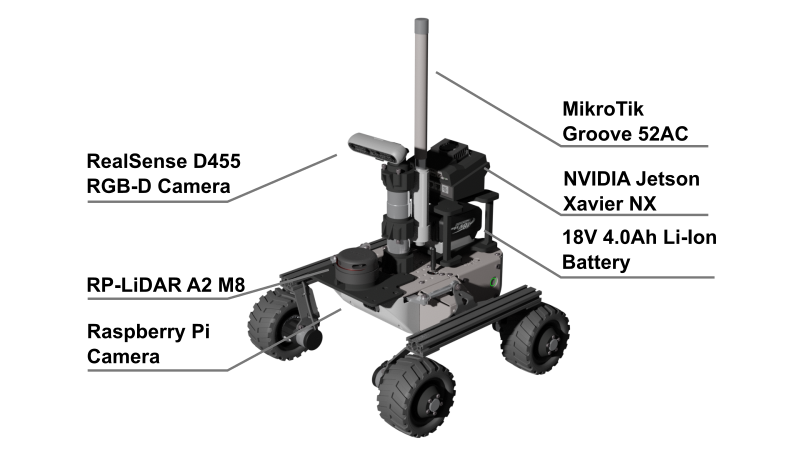}
\caption{Leo Rover setup used for \gls{realms2}}
\label{fig:leorover_realms2_small}
\end{figure}

The rovers are made of \gls{cots} components.
For a real lunar mission, space-grade hardware would be required.
The robotic base is a Leo Rover~\cite{leorover_tech}, designed as a space-rover analogue platform.
It features a native \gls{ros2} implementation running on a Raspberry Pi 4 board.
The upgraded robots for \gls{realms2} feature many improvements, as visible in Fig.~\ref{fig:leorover_realms2_small}.
To ensure complete mapping capabilities, each robot is equipped with an RPLIDAR~A2M8 along with a RealSense D455 RGB-D camera.
The sensors allow the robots to perceive their surroundings and provide insightful data, such as colour images, for future analysis.

The robot is equipped with a more powerful processing board, the NVIDIA Jetson Xavier NX, to process this amount of data.
This board features a GPU, which ensures proper processing of the large amount of data provided by the sensors.
The second board, the Raspberry Pi 4, is only in charge of the robot's control.
This ensures independence between motor control and sensor processing so that the operator can always maintain control.

Each robot is equipped with a Mikrotik Groove 52ac router.
This device corresponds to an antenna and a router capable of running the \gls{hwmp}+ protocol.
This allows each robot to act as a node in the mesh network, distributing the load and allocating the bandwidth more efficiently.
Since every robot is active in the network, it can serve as a relay by optimising its position to extend the system's range.

Two different rechargeable batteries power the robots.
The first is the default built-in battery that powers the internal Raspberry Pi and the motor controller to power the motors.
The second battery is from a cordless drill and is used to power the Mikrotik antenna.
This router connects all the onboard computers with the Mikrotik antenna, the Jetson Xavier and, therefore, the 2D LiDAR and the IntelRealsense camera.

\subsubsection{Software}

\begin{figure} [htp]
\centering
\includegraphics[width=1\columnwidth]{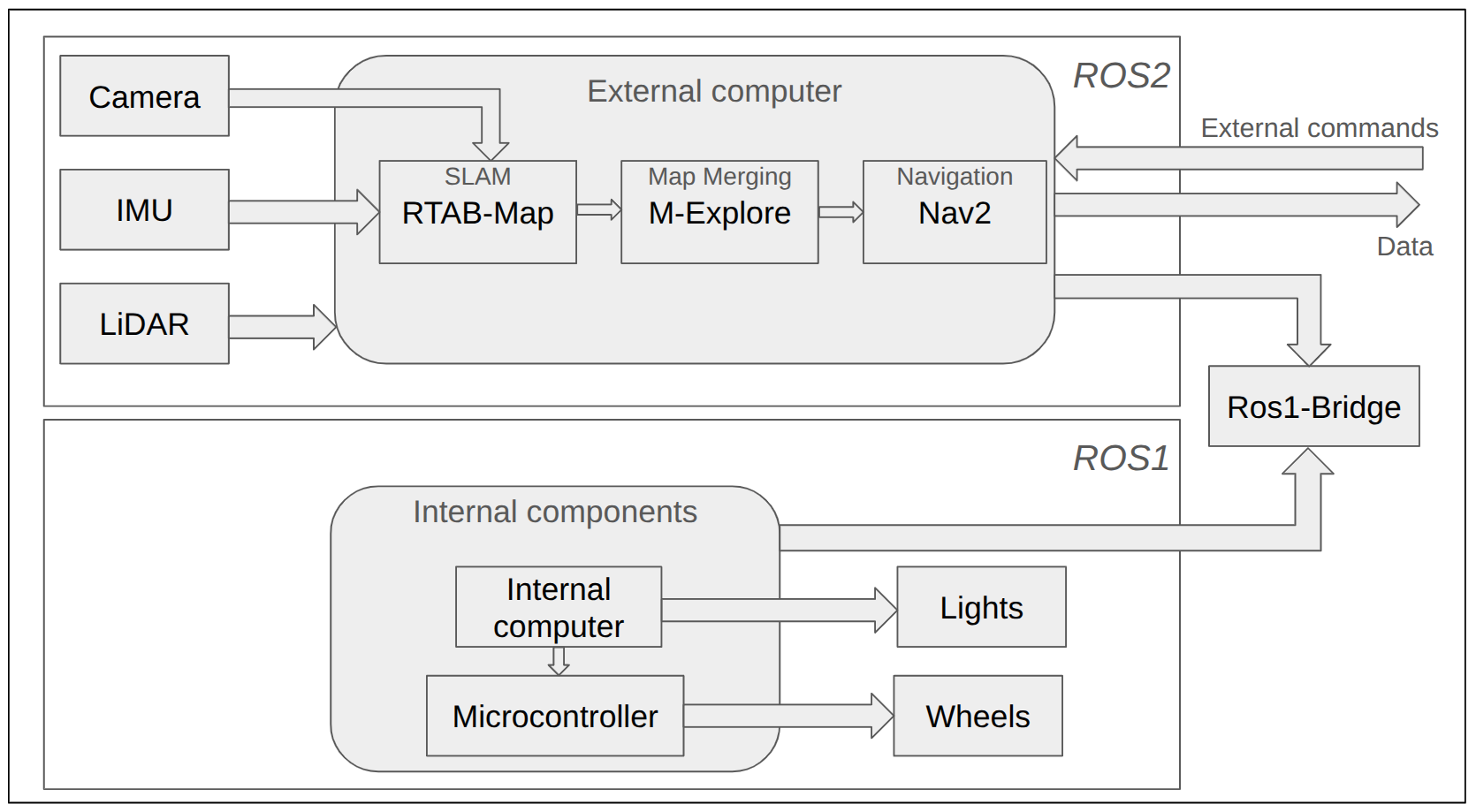}
\caption{\gls{realms2} global system overview}
\label{fig:realms2_overview}
\end{figure}

As shown in Figure \ref{fig:realms2_overview}, the lower part of the diagram illustrates the internal components and their interfaces for using the light and the wheels of the rover.
This internal system is managed by \gls{ros}.
Furthermore, the upper part of the diagram is composed of external components, including SLAM (RTAB-Map), map merging (M-Explore), and navigation (Nav2) capabilities, while ensuring communication with the operator.
The external system is managed by the \gls{ros2}.
A \gls{ros}-\gls{ros2} bridge makes the communication between \gls{ros} and \gls{ros2} data possible.

\gls{ros2} is the successor to ROS, addressing many of its limitations by incorporating modern software architecture principles.
By design, ROS is a fully centralised system relying on a master node, and making it decentralised would require using external packages, increasing the risk for potential issues \cite{REALMS1_2023}.
\gls{ros2} relies on the \gls{dds} standard \cite{Maruyama_Kato_Azumi_2016} for inter-node communication, which improves its real-time communication capabilities, security, and scalability.

However, \gls{ros2} still has some deficiencies compared to ROS, especially in terms of a more expensive computational load and data overhead due to the Quality of Service (QoS) policy, which ensures the quality of the messages \cite{Exploring_perf_ros2}.

The entire navigation stack, which consists of two main components, is implemented on the Nvidia Jetson Xavier.
A \gls{vslam} software, in charge of the mapping and localisation of the robot in its environment.
The \gls{vslam} software used is \gls{rtabmap}~\cite{LabbeMathieu2019Raao} allowing proper sensor merging at a loop closure.
The second component is the autonomous navigation software.
Once the robot knows its surroundings and position, it needs to know how to reach its next goal.
The software suite nav2 \cite{macenski2020marathon2} includes all of the sub-elements to convert any goal received by the robot to a given set of commands.
To improve the portability and streamline the startup sequence, the software is implemented in Docker containers.
This simplifies the deployment of the entire architecture to a new robotic platform while \gls{ros2} ensures interoperability.
Additionally, Docker containers can be configured to start immediately at the boot process of the robot, shortening the starting sequence.

A simulated communication delay of two seconds roundtrip time will affect the communication between the lander and the ground station.
This challenge needs to be addressed by the \gls{ros2} message system. An additional \gls{ros2} package is implemented, the map merging package \cite{map_merge_2021}, in order to simplify the user interface, merging all the different maps as one.
However, it requires each map to have at least 20\% of overlap.

\subsection{Lander}

The addition of a Lunar lander is an improvement to REALMS.
It consists of an Intel NUC, Core-i7 CPU, 32GB of RAM, a camera, and LiDAR on a static platform.
The primary usage of the lander is to act as a network gateway between the ground station and the \gls{mrs}.
This recreates a setup similar to a real mission, where a lander would be the only device with an antenna powerful enough to communicate with Earth.

Additionally, the embedded computer in the lander can act as an edge computing device, providing more computing power than the robots.
As a result, the lander can handle global tasks such as merging maps and backing up data.
Finally, the lander has a LiDAR and a camera to provide the robot operators with first insights into the environment.
With this information, the operators can make an exploration plan to optimize the mission time and focus on valuable resources.

\subsection{User Interface}\label{sub:user_interface}

The operator controls the rovers using a front-end system using a \gls{gui}.
A back-end system forwards the commands and handles the communication with the different rovers.

\subsubsection{Frontend}

Foxglove Studio~\cite{foxglove_2021} with a custom panel is used as frontend.
\gls{realms} uses RVIZ as a user interface.
It is the default visualisation software for \gls{ros}; it integrates well in the \gls{ros} ecosystem and is known by most users in the robotics community.
However, Foxglove provides some benefits, such as a more intuitive interface for creating custom panels.
The \gls{gui} allows the use of panels for different types of information and tabs to group panels together to provide a better overview to the operator.

\begin{figure} [htp]
\centering
\includegraphics[width=0.9\columnwidth]{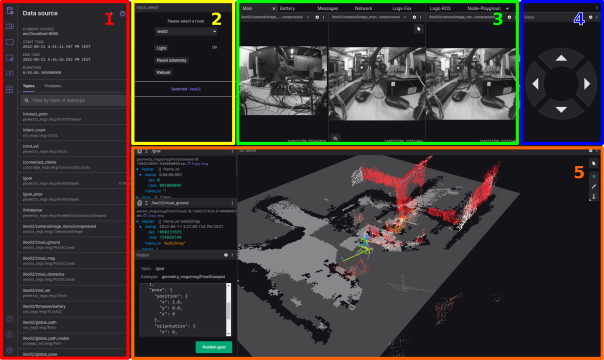}
\caption{Foxglove Studio - Main interface}
\label{fig:foxglove_studio_main}
\end{figure}

Fig.~\ref{fig:foxglove_studio_main} shows the frontend of \gls{realms2} using Foxglove.
The first panel (red outline) shows the panel properties and enables the adjustment of the settings for the selected panel.
The second panel (yellow outline) shows the custom robot selection panel, which allows the operator to choose which rover to send the commands.
This panel is robot-agnostic and scans the network for all available robots by filtering them through their namespaces.
The custom panel continuously looks for new namespaces in the system and adds them to the selection options.
The backend will use the selected robot name as a namespace to forward all the commands to the respective rover.
It also contains buttons to turn on or off the lights of the rover, reset the odometry or reboot the rover.
The third panel (green outline) shows the camera views of each of the three rovers.
The operator can supervise all rovers at the same time and interact if necessary.
The fourth panel (blue outline) contains a small teleoperation panel that allows the movement of the selected rover through the \gls{gui}.
The fifth panel (orange outline) shows the 3D view of the interface, displaying the 3D poses of the robots, the 2D projections of their respective maps, and the 3D point clouds of each rover.
The map merging of \gls{realms2} allows the user to see each of the three maps in a single viewport.
The operator can send autonomous navigation goals by clicking at a position in the 3D viewport, and the backend will forward the waypoint to the respective rover.
The use of Foxglove with the custom panel and the backend system enhances the scalability and usability of the system and makes it easier for a single operator to control multiple rovers.

\subsubsection{Backend}

The main function of the backend is to forward information from the rovers to the frontend and from the frontend back to the selected rover.
\gls{realms} requires at least one operator per robot.
This means that two operators are required to control two rovers.
During the \gls{esa}-\gls{esric} Challenge, the number of operators in the control room is limited to five people.
\gls{realms2} implements a back-end system to handle multiple robots through a single interface.
This allows the system to be more scalable.

The backend runs on the ground station computer and supports six features from the frontend:
rover selection, teleoperation, headlight control, odometry reset, rover reboot, and network monitoring.
The operator needs to select the rover to be controlled in the frontend.
The backend takes this parameter and forwards all the commands from the frontend to the selected robot by using the robot's name as the namespace for the commands.
If no rover has been selected, the namespace cannot be derived, and the backend does not send any commands to avoid unintended behaviour.

The primary function of \gls{realms2} is to send instructions to move the rovers.
These commands can serve as position goals for autonomous navigation or as teleoperation instructions for manually controlling the rovers.
In addition to forwarding commands to the selected robots, the backend monitors them, collects data and calculates bandwidth usage.
This monitoring method does not create additional overhead on the network.


\section{Subsystem Tests}
\label{sec:experiments}

In order to qualify and validate the \gls{realms2} architecture, we conducted experiments in the two main additions to \gls{realms2}: The network architecture and the map merging capabilities.

\subsection{Network Testing and Evaluation}

\gls{realms2} presents one of the first applications of a mesh network architecture for a \gls*{mrs} that can be used in space.
Due to this novel approach, the network capacities needed to be evaluated in order to develop the proper operational protocol for the missions. 
The three interesting points to quantify are:

\begin{itemize}
    \item The maximum connectivity range between two robots.
    \item The influence of a relay robot between the operator and a teleoperated robot
    \item The impact of the mesh network on the \gls{ros2} communication architecture
\end{itemize}

\subsubsection{Maximum Communication Range}

To evaluate the maximum range of the communication system using mesh routers, a robot was moved further away from the operator.
The experimental setup for this is shown in the upper half of Fig.~\ref{fig:range_experiment}.
The robot increases the distance to the operator until it loses connection. 
The distance measurement between the robot and the operator results in an estimated maximum range of 220 meters.
The chosen environment featured a direct \gls{los} scenario, with no direct obstruction, such as buildings, rocks or dense forests.
A space exploration mission is not expected to have a high density of obstacles. 
For a short-range mission, there is no \gls{los} that could be introduced by dunes or craters.
According to \cite{wright_nasa_2010}, a crater would need to have a diameter exceeding 5 km, and sometimes up to 20 km, to be deep enough for its rim to act as an obstacle.
It is worth noticing that the experimental site corresponds to an urban environment with some trees and moving obstacles. 
Therefore, the interferences can be considered more impactful than expected in a lunar scenario. 
As a result, the estimated value is a good baseline for the maximum range since the actual range would be higher in a real setup.

\subsubsection{Mesh Network Routing Capability}

\begin{figure} [htp]
\centering
\includegraphics[width=0.8\columnwidth]{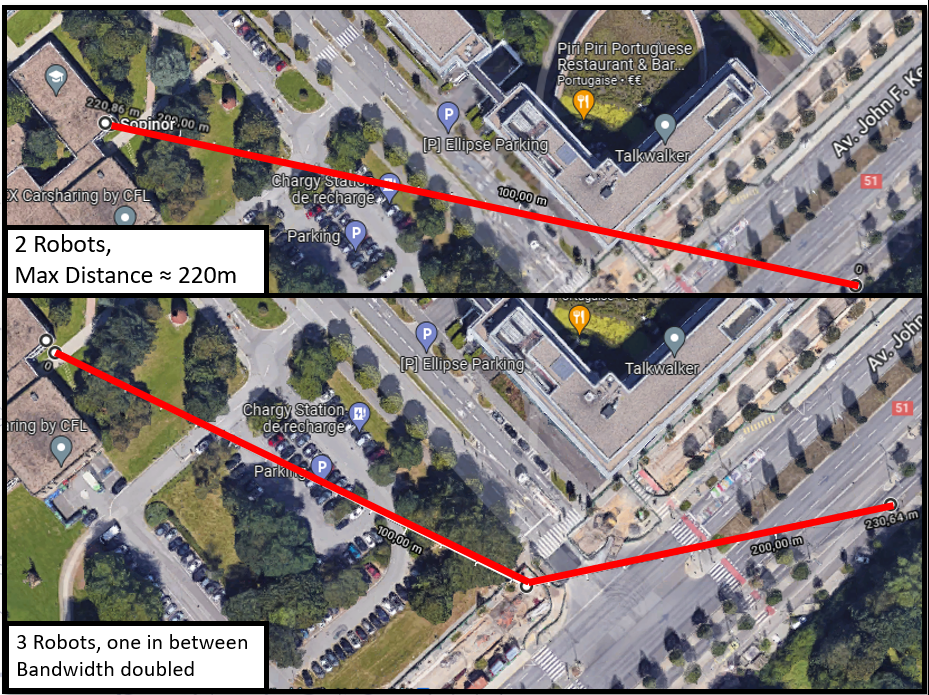}
\caption{Two graphs displaying the positions of the robots during the network evaluation. On the top, only two robots are used to measure the range. On the bottom one, a relay is placed in the middle, ensuring a better bandwidth}
\label{fig:range_experiment}
\end{figure}

To verify the capabilities of the mesh network to relay communication, the robot has been placed at the edge of the communication reach of 220~m to evaluate the network capacity.
At this distance, a ping command or teleoperation is still feasible, but the communication is too limited for visualisation.
A second rover is placed at 130~m away of the operator and 100~m of the teleoperated rover, as visualised in the lower half of Fig.~\ref{fig:range_experiment}.
In this situation, it is possible to receive images from the teleoperated rover, and the evaluated bandwidth capacity would increase between 2 to 10 times.

This additional knowledge provides strategies for operating the robots during the mission.
To ensure proper coverage, one of the robots should remain available to act as a relay and should not be operated on for more complicated tasks.

\subsubsection{Integration test of ROS 2 in a mesh network}

The mesh network needs to support \gls{ros2} messages through relay nodes.
The experiments highlight the differences in terms of network use between \gls{ros2} and its predecessor, ROS.
Setting ROS for multi-robot usage in a decentralised way required some workaround relying on installing a non-official package.
\gls{ros2} offers a more straightforward approach, providing native support for \gls{mrs}.
However, \gls{ros2} also comes with trade-offs.
The experiments revealed that the \gls{ros2} network architecture introduces additional overhead on the message sent, leading to more bandwidth usage by each process.
When testing the communication through the mesh network while simulating a communication delay, the messages were successfully transmitted using \gls{ros2}.
In REALMS, the communication delay prevented the rovers from connecting to the ROS master on the ground station.
Consequently, the system was modified to provide a ROS master for each robot and ground station computer.
This multi-master approach would rely on a non-standard approach, adding potential points of failure and reducing the scalability of the system.
In \gls{ros2}, this additional layer is not necessary to handle the delay.

\subsection{Mapping System Testing and Evaluation}

The mapping of the system is done using \gls{rtabmap}~\cite{LabbeMathieu2019Raao}.
Each robot creates a local map.
These local maps are then merged into a single global map using the \textit{m-explore} package~\cite{map_merge_2021}.

\subsubsection{Mapping the lunar surface}

The mapping system is based on REALMS~\cite{REALMS1_2023}.
It proved to successfully handle the extreme lighting conditions of a simulated lunar surface.
Furthermore, lab experiments show that mapping of unstructured terrain at a speed of \(0.05\ m/s\) is possible without loss of odometry.
In the event of odometry loss, the mapping system provides an interface to reset the odometry of the robot to a known position, which can be recovered from the last known position of the robot. In the worst-case scenario, a new map needs to be created and added to the global map through the map merging system.

\subsubsection{Online Map Merging}

\begin{figure} [htp]
\centering
\includegraphics[width=0.6\columnwidth]{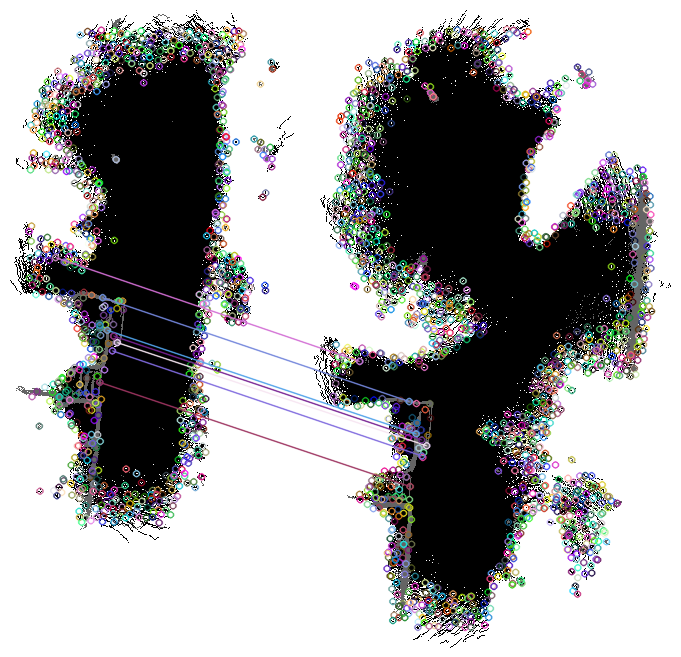}
\caption{Map merging - matches between two maps covering the same area}
\label{fig:map_mergin_matches}
\end{figure}

To test the map merging capabilities of \gls{realms2}, two rovers have been placed in an outdoor environment.
They start next to each other and are controlled through teleoperation to map the environment using \gls{vslam}.
At the start, none of the rovers is aware of the position of the other rovers.
After mapping the environment with a large overlap of the mapped area, the two maps are matched against each other using a map merging algorithm.
Fig.~\ref{fig:map_mergin_matches} shows the two maps and how their features are matched.
During this experiment, the two rovers are controlled using the user interface discussed in subsection \ref{sub:user_interface}.
The operator could see the camera streams of both rovers and the maps generated by both rovers.

\section{ESA-ESRIC Challenge}
\label{sec:challenge}
The final round of the \gls{esa}-\gls{esric} Challenge can be considered as the field test of \gls{realms2}.

\begin{figure} [htp]
\centering
\includegraphics[width=0.8\columnwidth]{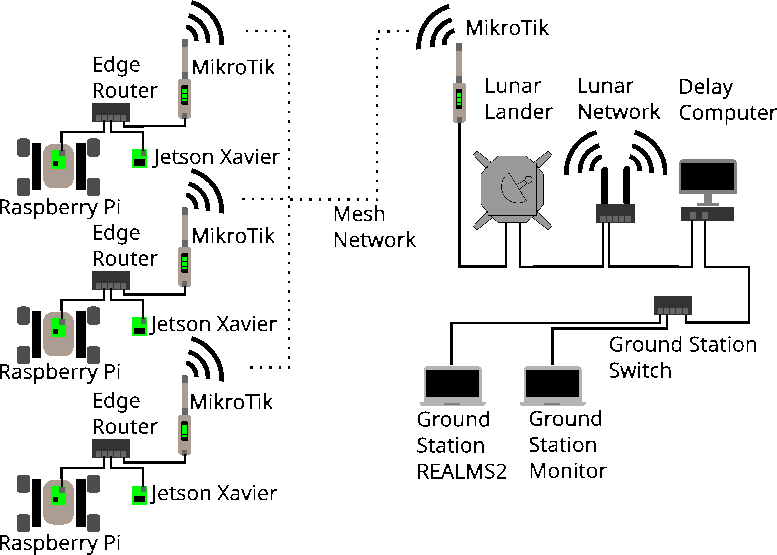}
\caption{Overview of the REALMS2 Network Architecture}
\label{fig:realms2_network_architecture}
\end{figure}

The challenge took place in an indoor environment in Esch-Belval, Luxembourg.
It featured an exploration zone of 50~m by 36~m that was unknown and not accessible to the operators before the challenge.
The area, Fig.~\ref{fig:ESA_ESRIC_SRC_second_field_test_arena},  featured various assets, such as a lander, craters, boulders, potential resources, and a floor made of finely grinded basalt.
The control room was connected to the lander through a network delay simulator as visible in Fig.~\ref{fig:realms2_network_architecture}.
The goal was to explore the area and identify valuable resources during the 4-hour run.

While the \gls{realms2} rovers were responsible for the mapping of the environment, the robot from \gls{sas} analysed the rocks using spectroscopy. 
The Leo Rovers were used to look for potential resources and provide an overview of the terrain.
When starting the mission, the autonomous navigation system of \gls{realms2} failed.
To continue the mission, two operators manually controlled two of the rovers through teleoperation.
During the challenge, the organisers requested that one rover be shut down to simulate a system failure so that they could verify resilience to unexpected events.
After this, the third \gls{realms2} rover was used to replace the rover with the simulated failure.
The lander represented the communication gateway between the rovers and the ground station.
During simulated communication blackouts, teleoperation was not possible. Therefore, each rover sent data to the lander to make efficient use of the bandwidth.


\section{Results}
\label{sec:results}

\begin{figure} [htp]
\centering
\includegraphics[width=0.99\columnwidth]{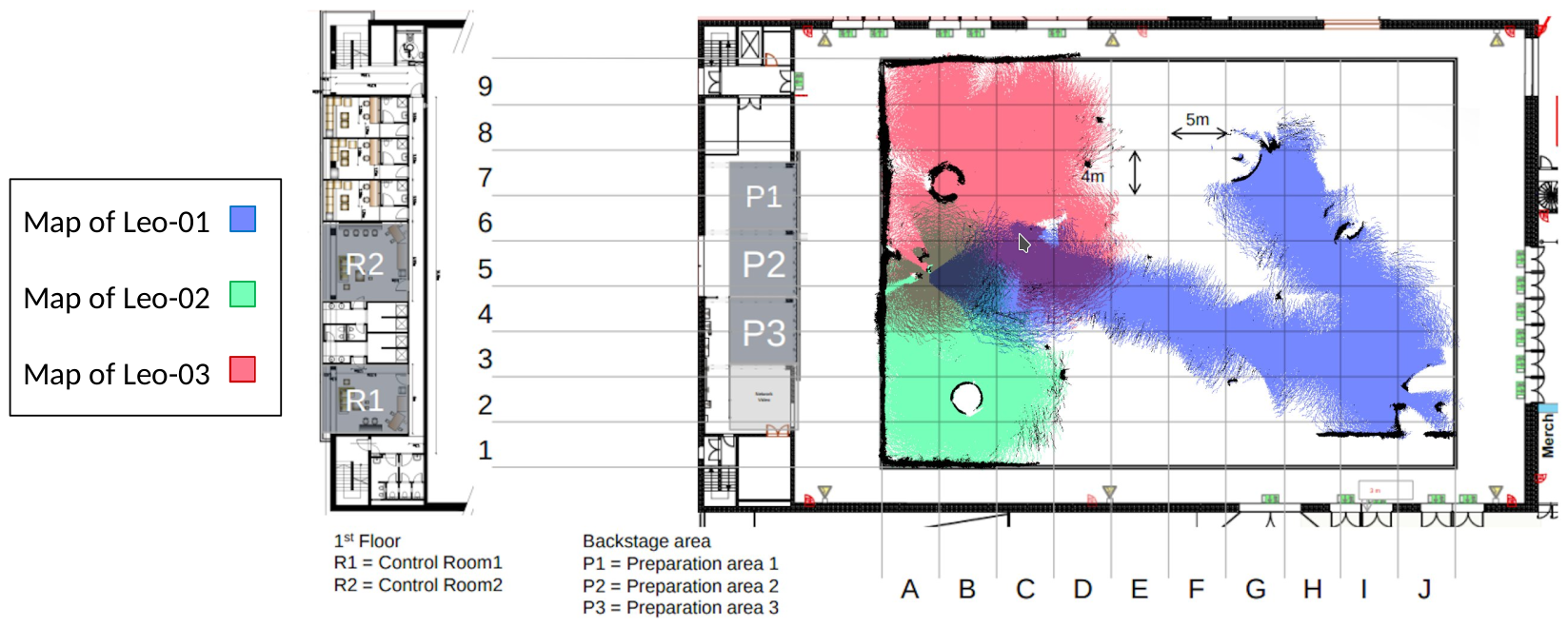}
\caption{Area Mapped by \gls{realms2} during the ESA-ESRIC Space Resources Challenge}
\label{fig:ESA_ESRIC_SRC_mapped_area}
\end{figure}

During the \gls{esa}-\gls{esric} Space Resource Challenge, \gls{realms2} was capable of mapping around \(60\%\) of the total surface of \(1800\ m^ 2\).
The three scouting robots encountered communication delays, blackouts, and a planned partial system failure to simulate the conditions of a real lunar mission.
Fig.~\ref{fig:ESA_ESRIC_SRC_mapped_area} shows the mapped area that highlights the contribution of each rover in blue, green, and red.
Together, the rovers covered about 60\% of the area.
This coverage is slightly below the expected results, as the performance was severely impacted by mission-specific conditions, including restricted operator interaction time, failure injection, and the need to revert to manual teleoperation after the autonomous navigation stack failed to execute planned paths.
Notably, this situation highlighted the \textbf{resilience} of the system, able to continue performing its tasks despite the various difficulties.

A minor scale issue can be observed in the blue part of the map in the bottom right corner.
A possible explanation is the scale estimation drift introduced by the slightly higher movement speed of this rover when traversing the terrain.
The map clearly shows two small craters on the left, a large crater on the bottom right, and several large rocks.
Due to the sparsity of the map features, the map merging algorithm was unable to calculate the relative transformation.
The rovers were controlled using teleoperation as the autonomous navigation stack failed to plan paths to the target locations of the rovers.

\section{Discussion}
\label{sec:discussion}

The \gls{realms2} system shows some limitations during laboratory experiments and field tests.
First, the map merging system only works in feature-rich environments, as it uses geometrical features in the map to find correspondences.
A system based on rich image data or scene semantics could reduce the dependence on geometric map features and reduce the overlap required to merge the maps.
Second, odometry loss is not detected automatically and requires intervention from the operator.
An automatic detection system followed by a predefined recovery behaviour could reduce the impact of temporary odometry loss.
Third, the autonomous navigation stack needs to be more robust.
If all three rovers were autonomously driving during the entire challenge, even more coverage could have been achieved.


\section{Conclusion}
\label{sec:conclusion}
This work presents a complete architecture of a distributed \gls{mrs} based on \gls{ros2}.
The SLAM capabilities of the system were enhanced, and improved resilience was demonstrated through the use of a mesh network.
In this setup, each robot acts as an independent relay for data and commands.
The use of Docker allows for an increase in modularity and thus improves the scalability of the system.
In addition, this approach increases the efficiency of the exploration, allowing multiple rovers to cover a larger area. 
This architecture has been tested in a lunar analogue facility, the LunaLab, and deployed in an expansive lunar environment created for the \gls{esa}-\gls{esric} Space Resource Challenge.
During the final round of the challenge, \gls{realms2} successfully mapped around \(60\%\) of the environment under realistic constraints, including induced communication blackouts and a forced system failure.
Despite the fallback to manual teleoperation, the system maintained continuous operation, highlighting its robustness and capacity to adapt to degraded conditions.


\section*{Acknowledgment}
The authors thank Space Application Services and their team in Belgium with special thanks to Jeremi Gancet, Fabio Polisano and Matteo de Benedetti for making the participation in the ESA-ESRIC Space Resources Challenge possible. The authors also want to thank ESA and ESRIC with special thanks to Massimo Sabatini, Franziska Zaunig, Thomas Krueger and Bob Lamboray for organising this event and hosting all team members in Esch-sur-Alzette, Luxembourg during the challenge. The authors thank Prof. Kazuya Yoshida from the Space Robotics Lab at the Tohoku University.

\section*{References}
\printbibliography[heading=none]

\end{document}